\documentclass[letterpaper, 10 pt, conference]{ieeeconf}  

\IEEEoverridecommandlockouts                              

\usepackage{graphicx} 
\usepackage{amsmath} 
\usepackage{amssymb}  
\usepackage{esvect}
\usepackage{changes}
\usepackage{float}
\usepackage{bm}
\usepackage{algorithm}
\usepackage{algorithmic}
\usepackage[caption=false, font=footnotesize]{subfig}
\def\realset{ \mathbb{R} }
\renewcommand{\baselinestretch}{0.994}

\title{\LARGE \bf
Autonomous and Teleoperation Control of a Drawing Robot Avatar}

\author{Lingyun Chen$^{1,2}$, Abdeldjallil Naceri$^{1}$, Abdalla Swikir$^{1,2,3}$, Sandra Hirche$^{4}$, and Sami Haddadin$^{1,2}$
\thanks{Funded by the German Research Foundation (DFG, Deutsche Forschungsgemeinschaft) as part of Germany’s Excellence Strategy – EXC 2050/1 – Project ID 390696704 – Cluster of Excellence “Centre for Tactile Internet with Human-in-the-Loop” (CeTI) of Technische Universität Dresden. The authors acknowledge the financial support by the Federal Ministry of Education and Research of Germany (BMBF) in the programme of "Souverän. Digital. Vernetzt." Joint project 6G-life, project identification number 16KISK002 and the funding of the Lighthouse Initiative Geriatronics by LongLeif GaPa gGmbH (Project Y). Please note that S. Haddadin had a potential conflict of interest as a shareholder of Franka Emika GmbH. 
}
\thanks{$^1$ Authors are with Munich Institute of Robotics and Machine Intelligence, Technical University of Munich, Germany, $^2$ also with the Centre for Tactile Internet with Human-in-the-Loop (CeTI), $^3$ and with Department of Electrical and Electronic Engineering, Omar Al-Mukhtar University (OMU), Albaida, Libya. $^4$ Chair of Information-oriented Control (ITR), Technical University of Munich, Germany. {\tt\small {lingyun.chen}@tum.de.}}%
}

\usepackage[normalem]{ulem}                                        

\providecommand{\colorsout}[1]{\bgroup\markoverwith{\textcolor{#1}{\rule[0.5ex]{2pt}{0.4pt}}}\ULon} 

\colorlet{an}{magenta}
\colorlet{lc}{orange}
\colorlet{as}{red}
\colorlet{sh}{blue}

\begin{document}

\maketitle
\thispagestyle{empty}
\pagestyle{empty}

\begin{abstract}

A drawing robot avatar is a robotic system that allows for telepresence-based drawing, enabling users to remotely control a robotic arm and create drawings in real-time from a remote location. The proposed control framework aims to improve bimanual robot telepresence quality by reducing the user workload and required prior knowledge through the automation of secondary or auxiliary tasks. The introduced novel method calculates the near-optimal Cartesian end-effector pose in terms of visual feedback quality for the attached eye-to-hand camera with motion constraints in consideration. The effectiveness is demonstrated by conducting user studies of drawing reference shapes using the implemented robot avatar compared to stationary and teleoperated camera pose conditions. Our results demonstrate that the proposed control framework offers improved visual feedback quality and drawing performance. 



\end{abstract}

\vspace{-10pt}

	\section{Introduction}


Modern telerobotic systems showcase high proficiency in performing complex and dexterous tasks in remote environments. These systems improve efficiency and cost-effectiveness across various industries, allowing tasks to be carried out with greater precision and speed while reducing human safety risks. The concept of the robot avatar, where a remotely operated robot embodies a human or provides a remote presence, is an integral part of these advanced telerobotic systems. As control, sensor and communication technology advances further, telerobotic systems, particularly in fields like medicine, exploration, entertainment, and disaster response, are set to become increasingly significant, augmenting the relevance of the robot avatar concept.


Visual feedback is the most straightforward and easy-to-implement method of delivering sensory information from the remote. One persistent challenge is dealing with the issue of occlusion in the visual feedback. The cameras often encounter objects that obscure parts of the view. This occlusion can hinder the user's ability to perceive the environment accurately and comprehensively. Actively positioning cameras is one way to achieve this. Due to work overload and network latency, it is challenging for the local user to acquire the best camera viewpoints in the telepresence setting while having a primary task in parallel. Therefore, camera viewpoint automation can reduce the user's time and effort required for the primary task and potentially improve performance. 
Camera viewpoint automation can be a valuable tool in robotic artistic applications, particularly robotic drawing systems, which have received a lot of interest recently \cite{song2018artistic, liu2021robust, song2022tsp}. 

A contact-rich manipulation task such as drawing shares the same fundamental challenges as various industrial and service robotic tasks where the robot pushes or grasps objects while maintaining contact with object surfaces \cite{hogan1987stable}. 
The integration of shared autonomy in dexterous manipulation holds immense potential to transform the way we approach these tasks.
We presented a drawing robot avatar with dexterous tele-manipulation which provides a research opportunity into fields such as remote control, tactile sensing, trajectory planning and human-robot interfacing \cite{chen2021drawing}. 
By utilizing camera viewpoint automation, a level of shared autonomy can also be added to the drawing robot avatar.


In this paper, we present a framework that allows one to implement a bimanual robot avatar for drawing by the local user through remote manipulation. This can also be deployed to medical, industrial and service settings such as surgery robots \cite{pandya2014review},  KI.Fabrik project \cite{KI.Fabrik} and GARMI robot \cite{trobinger2021introducing}, allowing remote surgery, home-office for factory workers and improving assist service quality to elderly, respectively. 
On the bimanual robot side, the drawing robot uses unified force and impedance controller (UFIC) to accurately draw on the paper while maintaining the required force. The camera-mounted robot uses a novel autonomous camera pose generation method to acquire better visual feedback for the local user. 
On the user side, an intuitive interface is developed to teleoperate the robot avatar with shared autonomy. 

In summary, the major contributions of this paper are:

\begin{enumerate}
    \item We introduced a novel camera control method for autonomous pose updates to enhance visual feedback quality for a bimanual drawing robot avatar. The method was evaluated through a combination of quantitative and qualitative assessments.
    \item A user study was conducted to compare the performance of the proposed autonomous camera pose with stationary and teleoperated camera poses. 
\end{enumerate}

	\section{Related Work}

\subsection{Drawing Robot Avatar}




\begin{figure*}[t]
\begin{center}
\vspace{7pt}
\includegraphics[width=14cm]{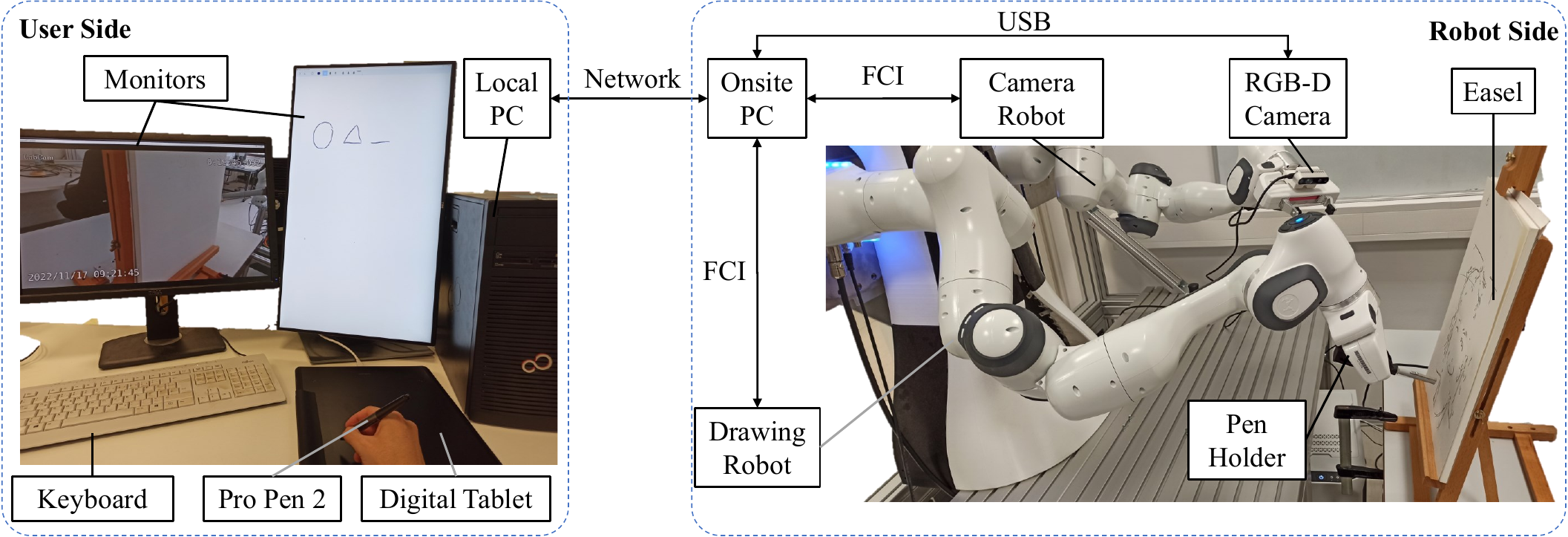}    
\vspace{-8pt}
\caption{Overview of the hardware setup. } 
\label{fig:robotsetup}
\end{center}
\vspace{-22pt}
\end{figure*}

Numerous implementations and research studies of robotic drawing systems have been done as a result of the quick development of robotic systems and control methods. There is a particular emphasis on utilizing a high-degree-of-freedom robotic system or manipulator for artistic drawing. 

Calinon et al. \cite{calinon2005humanoid} developed a robot capable of drawing artistic portraits. More recently, Song et al. \cite{song2018artistic} introduced a semi-autonomous robotic drawing system designed to produce pen artwork on any given surface. Liu et al. \cite{liu2021robust} presented a robotic system designed to autonomously create drawings on 3D surfaces. DrozBot, developed in \cite{low2022drozbot}, utilized an ergodic control algorithm to plan the robot paths for the drawing. These systems typically take images as input and recreate them using certain path planning and control algorithms on 2D or 3D surfaces. Our focus is more on empowering the robot avatar to draw based on telepresence. 

Attempts have also been made to incorporate teleoperation with robotic drawing systems. The PumaPaint project \cite{stein1998painting} enables online users to remotely draw paintings with camera image feedback. The implementation presented by \cite{wu2015improving} performs teleoperation experiments by drawing characters on balls. There are also recent attempts to utilize eye-tracking to teleoperate the drawing robot \cite{scalera2021performance, scalera2021human}.

In \cite{chen2021drawing}, we proposed a telepresence robotic system allowing artists to draw portraits via a 5G network. In our previous implementation, the camera was positioned at a predefined location which leads to occasional occlusion in the visual feedback due to the moving robot. 

By further developing it into the bimanual robot setup, the other robot is able to host the camera and provide different camera viewpoints which can potentially provide better visual feedback.

\subsection{Camera Viewpoint Automation}

Extensive research has been conducted related to camera viewpoint automation in the area of robotics and computer graphics. In general, camera position control is a specific case of path planning which makes it a PSPACE-hard task with exponentially increasing complexity in terms of the number of degrees of freedom \cite{christie2008camera}. One way in the telepresence setting is to allow users to change the camera viewpoint directly, such as the Vicarios VR interface introduced in \cite{naceri2021vicarios}. Though it can improve the situation awareness, the user still requires additional efforts to control the camera viewpoint. 

Various approaches are developed to solve the camera viewpoint automation challenge. Such as for path planning, user’s gaze \cite{zhu2011moving}, end-effector locations \cite{mudunuri2010autonomous}, object importance levels and the robot model can be used to generate an optimal trajectory for the camera with robot kinematics, dynamics, self-collision and visual obstruction constraints in consideration. Recently, Rakita et al. \cite{rakita2018autonomous, rakita2019remote} presented an approach aimed at enhancing the teleoperation of a robotic manipulation arm by consistently offering remote users an optimal perspective through the utilization of a second robot arm equipped with a camera-in-hand. We deployed a different occlusion optimization method based directly on the geometric mesh of the robot. 

The cognitive approaches have attracted research interest lately and use trained neural networks to decide the camera viewpoint based on offline demonstration \cite{rivas2019transferring, wagner2021learning}. In the telepresence setting, it's hard to get enough dataset to train the neural network model beforehand and the capability of generalizing to other applications remains questionable. 

In this paper, we aim to implement camera viewpoint automation within the drawing robot avatar to improve the telepresence quality and enhance the user experience. By automating the positioning of the camera viewpoint, the user can obtain a better view of the drawing surface and more effectively control the drawing process, leading to a more immersive telepresence experience.

\vspace{-6pt}
	\section{Drawing Robot Avatar}

\subsection{Hardware Setup}


The hardware setup for the drawing robot is shown in Fig. \ref{fig:robotsetup}, which consists of two 7-DoF Franka Emika robots \cite{haddadin2022franka} based on the GARMI design \cite{trobinger2021introducing} on the robot side. The left robot arm is equipped with an Intel Realsense depth camera D435 which is referred to as the camera robot. The camera can capture the image with a resolution of 1920*1080 in 30 frames per second. 
\begin{figure*}[!htb]
\begin{center}
\vspace{6pt}
\includegraphics[width=\textwidth]{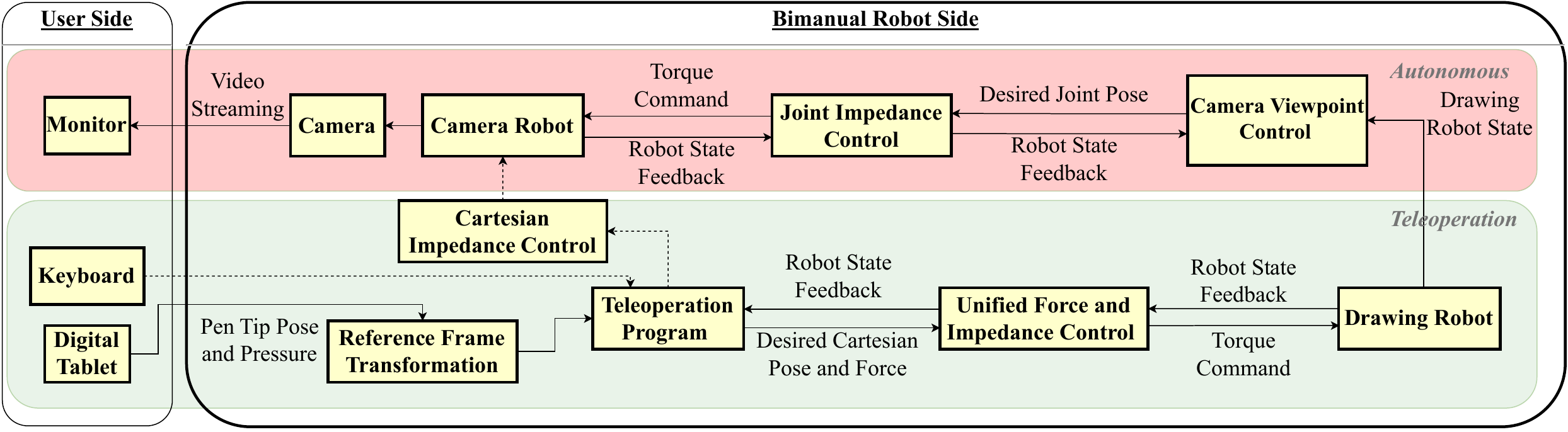}    
\vspace{-22pt}
\caption{Control framework of the proposed drawing robot avatar. (Dotted line arrows are only used in experiments as the keyboard control condition)}
\label{fig:ControlFramework}
\end{center}
\vspace{-20pt}
\end{figure*}
The right robot arm referred to as the drawing robot, grasps the pen holder with a spring inside to provide a buffer distance between the pen and the paper. 

Both robots are controlled using the Franka Control Interface (FCI) at 1 kHz. The developed local control software runs on a computer (Intel Core i5-12600K CPU @ 4.50GHz) installed with Ubuntu 20.04 LTS and real-time kernel (5.15.55-rt48).

On the user side, the setup consists of a PC, a digital tablet, a keyboard and monitors. The digital tablet is selected to ensure an intuitive way for the user to draw since it's already widely used among professional designers or artists. The digital tablet model is Wacom Intuos Pro Paper Edition L (PTH-860P), with a sampling frequency of 200 Hz. The tablet has a Wacom Pro Pen 2 which has 8192 levels of pressure sensitivity and is able to provide tilt information which can be used to control the pen orientation on the robot side. The keyboard allows the user to control the camera robot pose manually and the performance will be compared to the proposed autonomous camera pose generation algorithm. 

\subsection{Drawing Process}

The drawing process of the proposed bimanual robot includes a calibration phase and a teleoperation phase \cite{chen2021drawing}.

\subsubsection{Calibration Phase} 
The calibration phase happens before the teleoperation phase. The drawing robot touches the paper to acquire the drawing area location represented as a plane equation in the robot base frame. The process is semi-automatic. The robot is manually guided to a pose near the paper and then a force controller is used to apply the desired force on the paper. The plane equation can be derived through plane fitting after acquiring multiple end-effector Cartesian poses from the robot state. 

\subsubsection{Teleoperation Phase} To initiate safe teleoperation, a handshake process is required to avoid large discrepancies between the current drawing robot pose and the desired one from the user side. The handshake process synchronizes the drawing robot pose with the initial targeted pose based on the user input before the user command streaming takes over. During the handshake process, the camera robot also moves to the autonomously generated Cartesian pose. 



\begin{figure}[!ht]
\begin{center}
\vspace{-8pt}
\includegraphics[width=7cm]{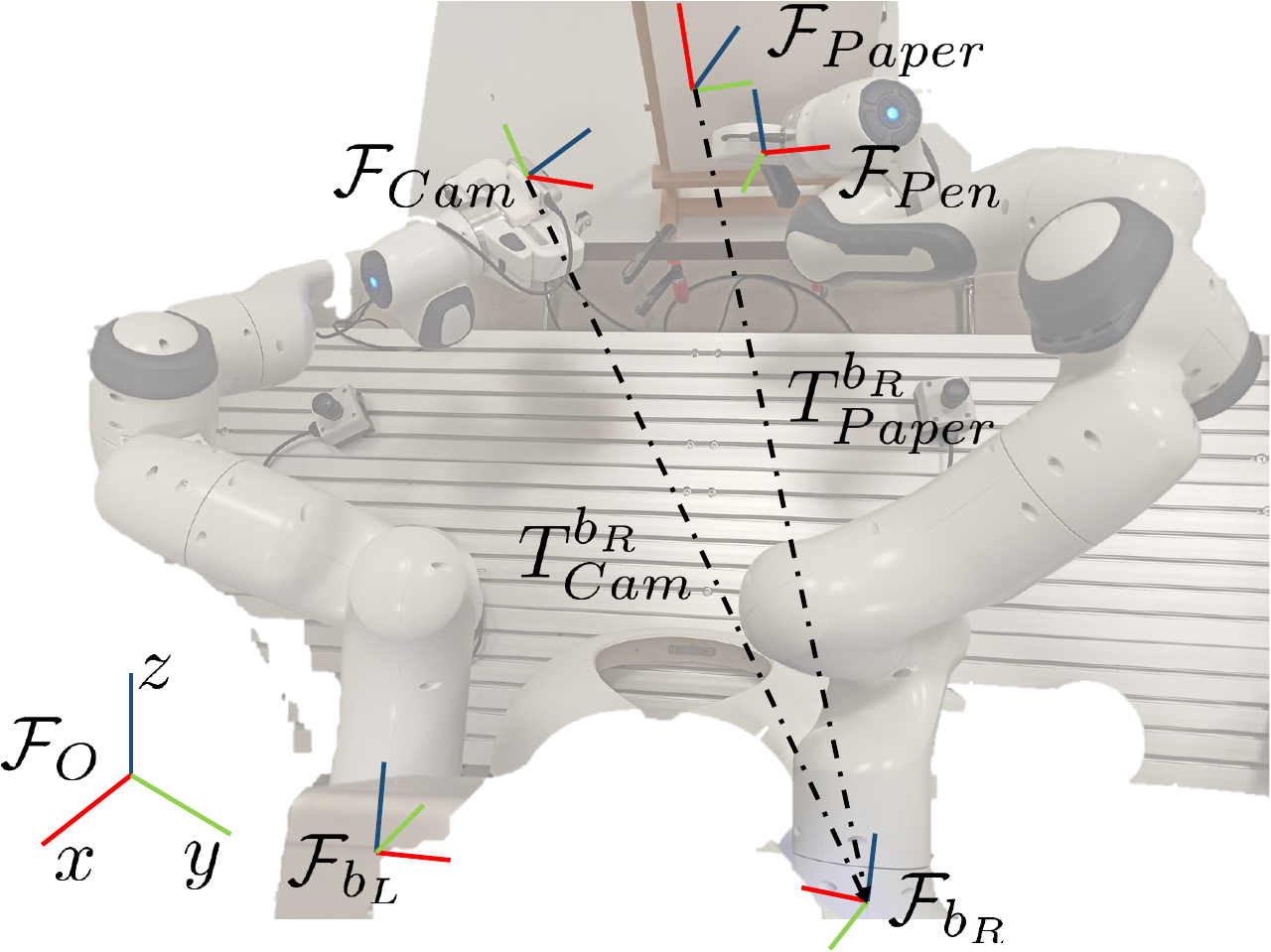}    
\vspace{-12pt}
\caption{Reference frames and spatial transform notations} 
\label{fig:SpatialTrans}
\end{center}
\vspace{-25pt}
\end{figure}
\subsection{Control Framework}

The overall architecture of the control framework is shown in Fig. \ref{fig:ControlFramework}. Fig. \ref{fig:SpatialTrans} illustrates the reference frames and spatial transform notations. The camera frame of the camera robot $\mathcal{F}_{Cam}$, the end-effector (pen-tip) frame of the drawing robot $\mathcal{F}_{Pen}$, the paper frame $\mathcal{F}_{Paper}$ as well as the origin frame $\mathcal{F}_O$ are depicted. The transformation of  $\mathcal{F}_{Cam}$ in the drawing robot base frame $\mathcal{F}_{b_R}$ is denoted as $\boldsymbol{T}^{b_R}_{Cam}$. The transformation of $\mathcal{F}_{paper}$ in $\mathcal{F}_{b_R}$ is described as $\boldsymbol{T}^{b_R}_{Paper}$.

To provide a structured description of our control framework, we divide our approach into three main subsections, 1) direct Cartesian space teleoperation control which converts the user inputs from the tablet to desired drawing robot poses and applied forces, 2) UFIC which converts the desired drawing robot state to joint torque commands, and 3) autonomous Cartesian pose generation which controls the camera robot to achieve better viewpoint quality.

\subsubsection{Direct Cartesian Space Teleoperation Control}
\begin{figure}[!ht]
\vspace{-8pt}
\begin{center}
\includegraphics[width=7cm]{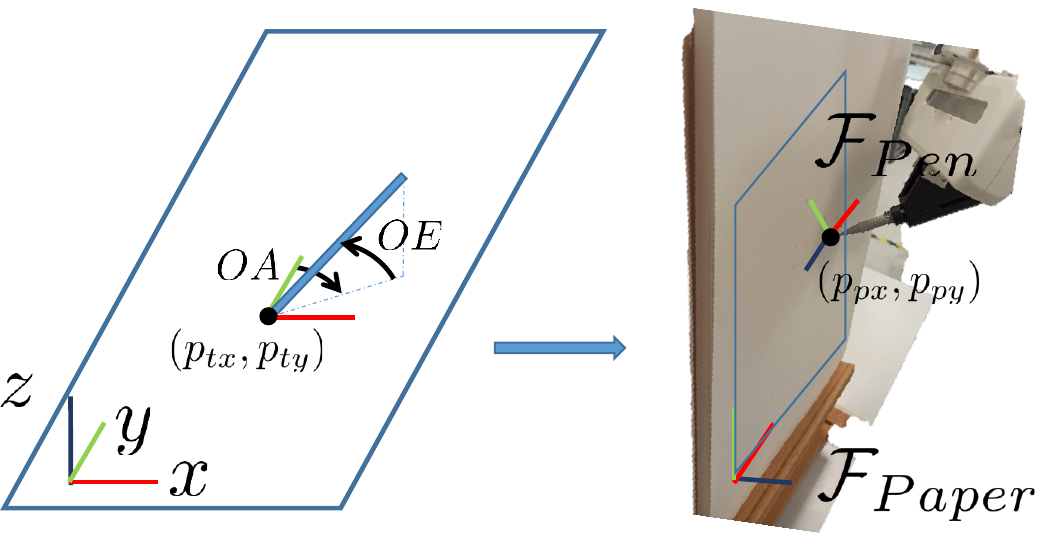}    
\vspace{-15pt}
\caption{Transformation from digital tablet pose to drawing robot pose} 
\label{fig:DirectTransform}
\end{center}
\vspace{-14pt}
\end{figure}

The digital tablet transmits the pen tip pose and pressure to the teleoperation program, as shown in Fig. \ref{fig:DirectTransform}. The pose includes the position defined by $x$ and $y$ coordinates $(p_{tx}, p_{ty})$, pressure $\mathcal{P}$, altitude angle $OE$ and azimuth angle $OA$. The coordinates and pressure are normalized to $[0, 1]$. The altitude angle represents the angle between the pen and the tablet measured in radians between $[0, \pi/2]$. The azimuth angle indicates the angular distance along the plane of writing to the location of the pen ranging from $[0, 2\pi]$.

\par The goal of teleoperation control is to get the robot's Cartesian pose $\boldsymbol{X}^{bR}_{Pen}$ according to the input from the digital tablet shown in Fig. \ref{fig:DirectTransform}. The transformation process is shown in Alg. 1. Where $\boldsymbol{R}_z$ and $\boldsymbol{R}_x$ are the rotation matrix around $z$ and $x$ axis, respectively.

\par 
To ensure a minimum force required for drawing on the paper while avoiding excessive force applied to the paper and pen, a linear scale $F_{pen} = F_{min} + \mathcal{P}(F_{max}-F_{min})$ is utilized to obtain the target force $F_{pen}$ on the paper based on the digital tablet pressure reading. It should be noted that the force direction is perpendicular to the paper plane.
\vspace{-8pt}
\begin{algorithm}[htb]
    \label{alg1}
    \caption{Space transformation}
	\renewcommand{\algorithmicrequire}{\textbf{Input:}}
	\renewcommand{\algorithmicensure}{\textbf{Output:}}
	\begin{algorithmic}
	    \REQUIRE State of the pen:
     $p_{pan}^{paper}=(p_{tx},p_{ty},0)$, $OA$, $OE$,
     \newline Paper size: $p_{xmax}$,$p_{ymax}$, and transformation matrix for Robot in paper frame: $\boldsymbol{T}^{bR}_{paper}$
	    \ENSURE Robot target Cartesian pose $\boldsymbol{X}^{bR}_{Pen}$
        \STATE Calculate rotational transformation matrix for pen tip in the paper frame: $\boldsymbol{R}^{paper}_{pen}=\boldsymbol{R}_z(\pi/2-OA)\boldsymbol{R}_x(OE)$
        \STATE Calculate translation vector for pen-tip in the paper frame: $\boldsymbol{p}^{paper}_{pen} = (p_{xmax},p_{ymax},0) \odot (p_{tx},p_{ty},0)$
        \STATE The combined transformation matrix:
        \newline $\boldsymbol{T}^{paper}_{pen} \leftarrow \boldsymbol{p}^{paper}_{pen}, \boldsymbol{R}^{paper}_{pen}$
        \STATE $    \boldsymbol{X}^{b_R}_{Pen} = \boldsymbol{T}^{paper}_{pen} \boldsymbol{T}^{b_R}_{paper}$
	\end{algorithmic}
\end{algorithm}
\vspace{-8pt}

\subsubsection{Unified Force and Impedance Controller}
The desired pose $\boldsymbol{X}^{b_R}_{Pen}$ and force $F_{pen}$ become the input to the robot controller. In order to regulate the force applied to the paper by the drawing robot while following the input trajectory, the UFIC is used.


The classical impedance  and force control can be extended and unified into a single framework of unified force impedance control by composition of an impedance control torque $\bm{\tau}_{imp}$ with a force control torque $\bm{\tau}_{fc}$ \cite{shahriari2019power}, 
\begin{equation}
    \bm{\tau}_r = \bm{\tau}_{imp} + \bm{\tau}_{c}.
    \label{equ:UFIC}
\end{equation}
where $\boldsymbol{\tau}_r$ is the desired commanded torque to the robot.
The well-known impedance controller ensures that the robot follows a desired trajectory with the following control law,
\begin{equation}
\begin{aligned}
    \bm{\tau}_{imp} = & \boldsymbol{J}^{T}(\boldsymbol{q})(\boldsymbol{M}_{C}(\boldsymbol{q}) \ddot{\boldsymbol{X}}^{b_R}_{Pen} + \boldsymbol{C}_{C}(\boldsymbol{q}, \dot{\boldsymbol{q}}) \dot{\boldsymbol{X}}^{b_R}_{Pen} \\
    &+ \boldsymbol{K}_{x} \tilde{\boldsymbol{x}}+\boldsymbol{D}_{x} \dot{\tilde{\boldsymbol{x}}})
\end{aligned}
\end{equation}
where $\boldsymbol{q} \in \realset^{7}$ is the joint state, 
$\tilde{\boldsymbol{x}} =\boldsymbol{X}^{b_R}_{Pen}-\boldsymbol{x}$ is the tracking error in task space $\boldsymbol{x}$. $\boldsymbol{J}(q)$ denotes the Jacobian matrix, $\boldsymbol{K}_{x}$ and $\boldsymbol{D}_{x}$ are the stiffness and damping matrices, and $\boldsymbol{M}_{C}(\boldsymbol{q})$ and $\boldsymbol{C}_{C}(\boldsymbol{q}, \dot{\boldsymbol{q}})$ are the robot inertia matrix and Coriolis matrix in Cartesian space. 




The force control which regulate the interaction force $\boldsymbol{f}_{ext}$ to following a desired force profile $\boldsymbol{f}_{des}$ which can be constructed from $F_{pen}$ is given by
\begin{equation}
\begin{aligned}
    \boldsymbol{\tau}_{c} = & \boldsymbol{J}^T (\boldsymbol{q}) (\boldsymbol{f}_{des} + \boldsymbol{K}_{p} (\boldsymbol{f}_{des} - \boldsymbol{f}_{ext} ) + \boldsymbol{K}_{d} (\dot{\boldsymbol{f}}_{des} - \dot{\boldsymbol{f}}_{ext} ) \\ &  + \boldsymbol{K}_{i}  \boldsymbol{h}_i (\boldsymbol{f}_{ext}, t)),
\end{aligned}
\end{equation}
where $\boldsymbol{h}_i (\boldsymbol{f}_{ext}, t) := \int_{0}^{t} (\boldsymbol{f}_{des}(\sigma) - \boldsymbol{f}_{ext}(\sigma) ) d \sigma$. $\boldsymbol{K}_{p}$, $\boldsymbol{K}_{i}$ and $\boldsymbol{K}_{d}$ are the proportional, integral and derivative gains respectively.



The passivity of the closed-loop system is not preserved during environmental interaction. An energy tank can be further designed and augmented to ensure the passivity and thus stability of the system \cite{shahriari2019power}. 


\subsubsection{Autonomous Cartesian Pose Generation}
In order to realize camera pose generation automation, a
way to evaluate the quality of a specific camera viewpoint is required.
For the camera robot, a search-based method is used to find the target Cartesian pose $\boldsymbol{X}^{b_L}_{Cam}(t+1)$ for the next time step, as shown in Alg. 2. The time step size $\Delta t = 33 ms$ was chosen to be aligned with the camera frame time.
\vspace{-10pt}
\begin{algorithm}[!htb]
    \label{alg2}
    \caption{Generate possible Cartesian pose for next time step.}
	\renewcommand{\algorithmicrequire}{\textbf{Input:}}
	\renewcommand{\algorithmicensure}{\textbf{Output:}}
	\begin{algorithmic}
	    \REQUIRE Camera Robot Current State: $\boldsymbol{X}^{bL}_{Cam}(t)$, $\boldsymbol{\dot{X}}^{b_L}_{Cam}(t)$, \newline
         Cartestian motion limits: $\dot{p}_{maxt}$, $\dot{p}_{maxr}$, $\ddot{p}_{maxt}$ and $\ddot{p}_{maxr}$, and the number of samples desired in one direction: $n$.
	    \ENSURE Possible range for $\boldsymbol{X}^{b_L}_{Cam}(t+1)$.
        \STATE Calculate space resolution: ${S}_r=\ddot{p}_{maxt}/n$
        \STATE Sampling possible translation acceleration for each direction:
        $\{\ddot{p}_{trans}\} = \{0, \pm S_r, \pm 2S_r, ..., \pm n S_r\}, n \in \mathbb{N}$
        \STATE Calculate and limit rotation acceleration to focus the pen-tip to the center of camera. $\ddot{p}_{rotate}\in [-\ddot{p}_{maxr}, \ddot{p}_{maxr}]$
        \STATE Calculate the possible translation vector and rotation matrix:$\{\boldsymbol{T}_{trans}\} \leftarrow \{ \ddot{p}_{trans}\}$, $\{\boldsymbol{R}_{rotate}\} \leftarrow \{ \ddot{p}_{rotate}\}$
        \STATE Calculate the possible Cartesian acceleration:$\{\boldsymbol{\ddot{X}}_{Cam}^{bL \ast}\} \leftarrow \{\boldsymbol{R}_{rotate}\},\{\boldsymbol{T}_{trans}\}$
        \STATE Calculate the possible robot state for next time step:
        $\{\boldsymbol{X}^{bL}_{Cam}(t+1)\}=\boldsymbol{X}^{bL}_{Cam}(t)+ \boldsymbol{\dot{X}}^{bL}_{Cam}(t) \Delta t+$ \newline
        $\dfrac{1}{2}\{\boldsymbol{\ddot{X}}_{Cam}^{bL \ast}\}\Delta t^2$
	\end{algorithmic}
\end{algorithm}
\vspace{-10pt}

\begin{figure}[!ht]
\begin{center}
\includegraphics[width=8.0cm]{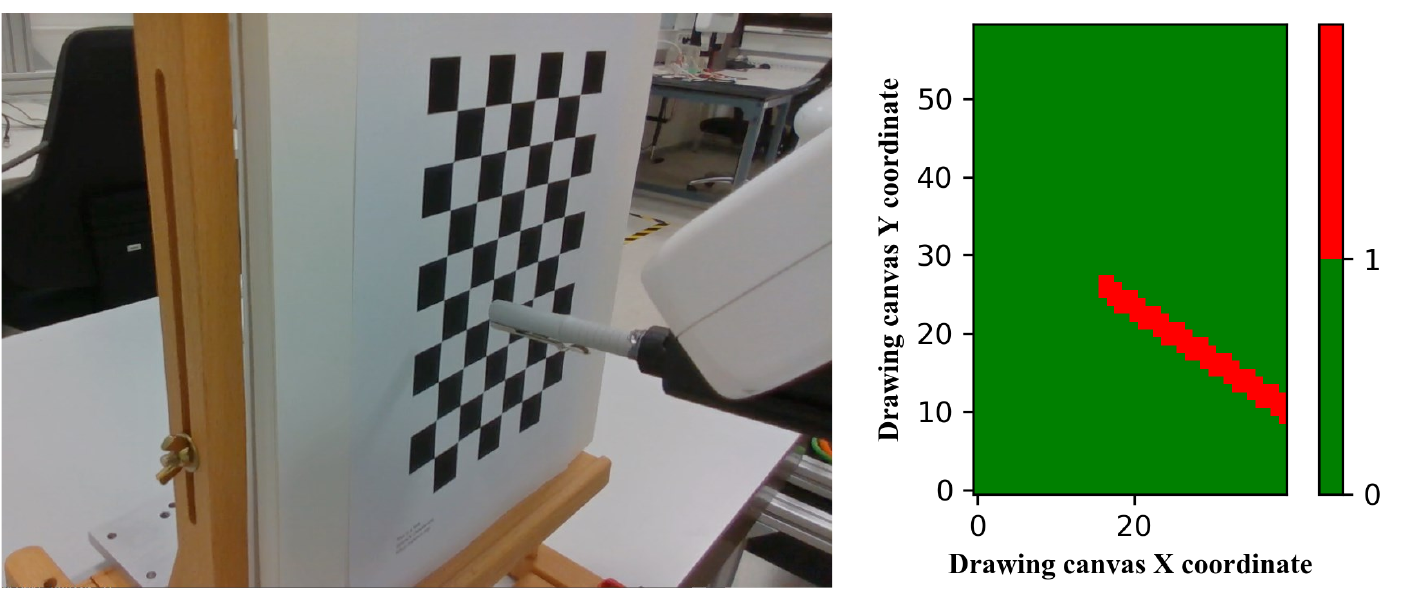}    
\vspace{-8pt}
\caption{Left: visual feedback. Right: Intersection detection results} 
\label{fig:IntersectionDetectionResults}
\end{center}
\end{figure}

The reachable Cartesian poses will be evaluated based on the viewpoint quality of the camera using Alg. 3, $g(i,j)$ is used to determine whether the sight-line between the camera and the point $\boldsymbol{X_{i,j}}$ is occlusion by the drawing robot. To improve computational speed, an AABB tree \cite{cgal:atw-aabb-22b} is constructed to enable rapid intersection detection. It is a hierarchical structure that divides space into hierarchical boxes to efficiently determine the relative positions and occlusion relationships between objects. In addition, the 3D mesh models are generated from a simplified robot 3D model to reduce the computation cost. The process of simplification involves reducing the number of vertices, edges and faces in the model, and removing internal geometry features that are not relevant to the viewpoint quality analysis being performed. Let $\boldsymbol{q}_{R}(t)$ denote the joint pose of the drawing robot at a time step $t\in \mathbb{R}_{\geq0}$. The joint pose $\boldsymbol{q}_{R}(t)$ generates a configuration of the robot body consisting of kinematic chain parameters and 3D mesh models $\{\Theta_{l}\}$, where $\Theta_{l}$ represents the 3D mesh model of the robot link number $l$. An intersection detection result is shown in Fig. \ref{fig:IntersectionDetectionResults}.

\vspace{-8pt}

\pagebreak
We assume that for remotely operated users, avoiding long-term occlusion of already painted areas can boost user confidence. Therefore, we introduce a weight function to consider the memory effects. For points that the drawing robot has painted, calculate the time since the camera last saw these points as $t_l$. If these points have been covered by the drawing robot for an extended period, the weight function $w(i,j)$ will decrease to encourage the robot to move to a position where it can see these points again. The memory parameter $c_m$ is a constant. Higher $c_m$ indicates a more dominant memory effect. $c_m$ is selected to be $0.18$ based on memory study in \cite{yakovlev2013s}. 

Finally, an analytical inverse kinematics method \cite{he2021analytical} is used to acquire the joint pose for the camera robot and pass it to the joint impedance controller. 
\vspace{-6pt}
\begin{algorithm}[htb]
    \label{alg3}
    \caption{Select the next target Cartesian pose}
	\renewcommand{\algorithmicrequire}{\textbf{Input:}}
	\renewcommand{\algorithmicensure}{\textbf{Output:}}
	\begin{algorithmic}
	    \REQUIRE Possible range for $\{\boldsymbol{X}^{b_L}_{Cam}(t+1)\}$, \newline
     the mesh model of the drawing robot at current time: $\{\Theta\}$, \newline
     the recorded past Cartesian poses of the drawing robot: $\{\Lambda\} = \{\boldsymbol{X}^{b_R}_{Pen}(t_{all}), t_{all}\in \left[0, t\right)\}$, \newline
     and sampling number for the drawing canvas: $S_x$,$S_y$.
	    \ENSURE The reachable Cartesian pose with the maximum viewpoint quality: $\boldsymbol{X}^{b_L}_{Cam}(t+1)$
        \STATE Convert the reachable Cartesian poses to the drawing robot frame:$\{\boldsymbol{X}^{b_R}_{Cam}(t+1)\} \leftarrow \{\boldsymbol{X}^{b_L}_{Cam}(t+1)\}$
        \FOR{$i \leq S_x$}
            \FOR{$j \leq S_y$}
                \IF{$\vv{\boldsymbol{X}^{b_R}_{Cam}\boldsymbol{X_{i,j}}} \cap \{\Theta\} \neq \varnothing$}
                    \STATE Set $g(i,j)=0$ 
                \ELSE
                    \STATE Set $g(i,j)=1$
                \ENDIF
                \IF{$\boldsymbol{X_{i,j}} \cap \{\Lambda\} \neq \varnothing$}
                    \STATE Set weight function $w(i,j)=2- e^{-c_m t_l}$
                \ELSE
                    \STATE Set weight function $w(i,j)=0.1$.
                \ENDIF
                \STATE Update the pose evaluation function: \newline
                $f(\boldsymbol{X}^{b_R}_{Cam})=f(\boldsymbol{X}^{b_R}_{Cam})+g(i,j)w(i,j)$
                \STATE $j=j+1$
            \ENDFOR
            \STATE $i=i+1$
        \ENDFOR
        \STATE Find the Cartesian pose with maximum viewpoint quality: $\boldsymbol{X}^{b_R}_{Cam} = argmax(f(\boldsymbol{X}^{b_R}_{Cam}))$
	\end{algorithmic}
\end{algorithm}
\vspace{-6pt}

\subsection{User Interface}

The user interface for the drawing robot avatar consists of three main parts, which are designed to provide the user with an intuitive and informative telepresence experience. The first part is visual feedback, which is provided through live video streaming of the camera using Real-time Transport Protocol (RTP) over User Datagram Protocol (UDP) with forward error correction (FEC) to reduce the latency while maintaining robustness to network packet loss. 


However, because video streaming requires additional time for encoding and decoding, it is not the fastest way to provide feedback to the user. Therefore, direct robot state feedback, which includes information about the contact state of the robot's end-effector with the paper, provides the user with low-latency information to react to. The robot's overall external force and end-effector pose are also displayed via real-time plotting along with the contact state. 

Finally, the user interface also includes a drawing input display, which displays the current tablet pen position and past drawing input.

	\section{Experiments}

To assess the effectiveness of the proposed robot telepresence system and the benefit of using camera viewpoint automation, the proposed system is evaluated qualitatively and quantitatively. Below, we present the details of our experimental procedures, conditions, hypotheses, analysis metrics, results and discussion.

\subsection{Experiment Procedure}
\begin{figure}[!ht]
\begin{center}
\includegraphics[width=4cm]{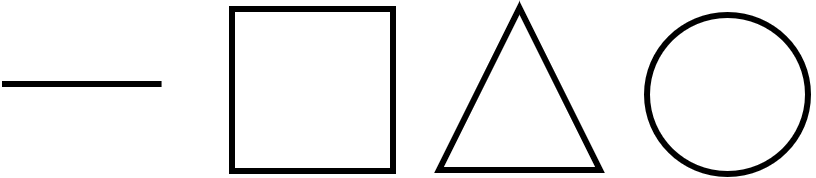}    
\vspace{-6pt}
\caption{Reference shapes.}
\label{fig:ExpReference}
\end{center}
\vspace{-20pt}
\end{figure}

Participants were trained to use the system for 5 minutes first with guidance. After that, 10 minutes were given to the user to get further familiar with the system. Following, participants were asked to draw reference shapes of a line, a square, a triangle and a circle shown in Fig.
\ref{fig:ExpReference} using the digital tablet. These reference shapes are selected due to the simplicity for the user to follow. 

\vspace{-7pt}

\subsection{Experiment Conditions}

For a comparative evaluation, the trials were performed in three different conditions: 

\begin{enumerate}
    \item \textit{Stationary camera pose (ST-Pose)}: The camera robot stays stationary at a predefined pose. 
    \item \textit{Teleoperated camera pose (TE-Pose)}: The camera robot is teleoperated by the user using the keyboard. Keyboard action modifies the Cartesian pose of the robot directly. There are twelve keys that control step increments and decrements on certain translational or rotational axes of the Cartesian pose.
    \item \textit{Autonomous camera pose generation (AU-Pose)}: The camera robot is controlled by the proposed autonomous camera pose generation framework. 
\end{enumerate}

Each participant conducted one trial for each condition. At the end of the experiment, participants filled in a post-experiment questionnaire, rating certain pre-determined properties of the different conditions on a 10-point Likert scale (1: bad and 10: good). 
The properties include mental demand, physical demand, drawing accuracy, motion continuity, visual feedback quality, and overall performance.

\begin{figure*}[ht]
\begin{center}
\includegraphics[width=\textwidth]{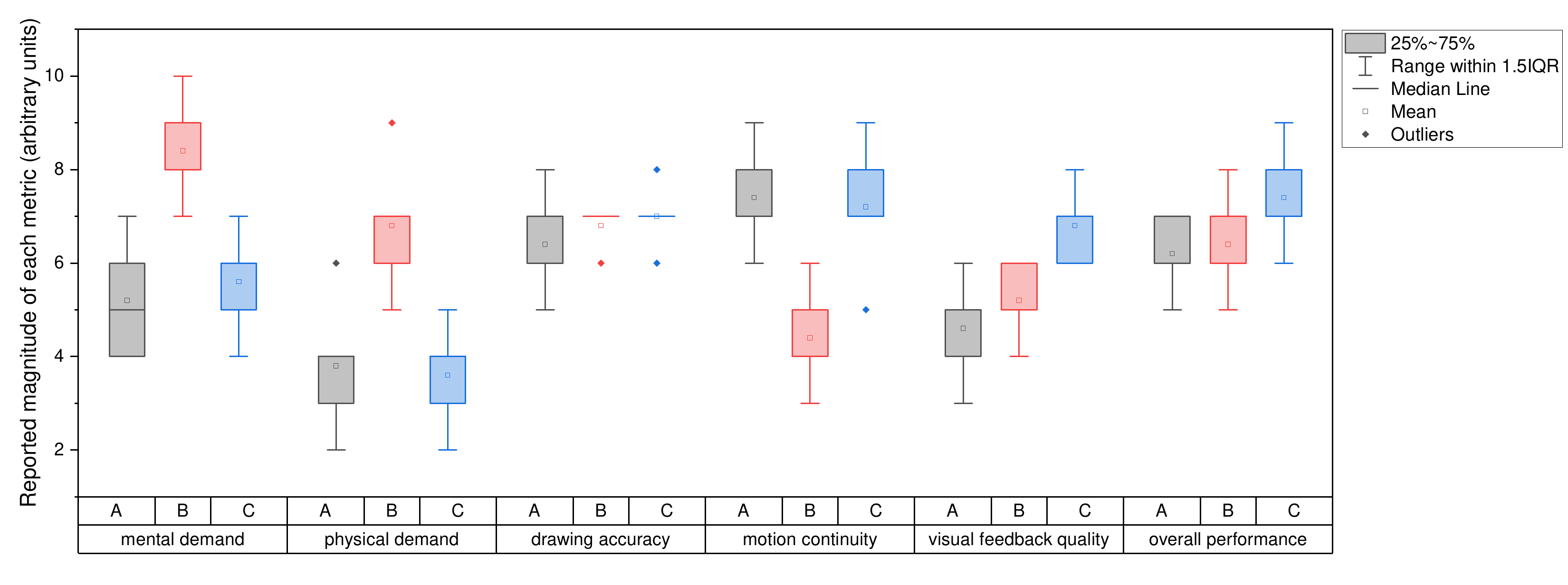}    
\vspace{-28pt}
\caption{Post-experiment questionnaire results based on three conditions. (A. ST-Pose. B. TE-Pose. C. AU-Pose)} 
\label{fig:boxchart}
\end{center}
\vspace{-30pt}
\end{figure*}

\subsection{Participants}

Five participants (age: 27 $\pm$ 3, two females) took part in our experiments. All participants were right-handed and had no prior interaction with the robot drawing avatar. 

\subsection{Hypotheses}

We have two hypotheses: $\boldsymbol{H_0}$: We predict that participant performance in the \textit{AU-Pose} condition would be similar to the baseline \textit{ST-Pose}, but better than the \textit{TE-Pose} condition in terms of task completion time. Indeed, we expect participants to be slower when manually teleoperating the camera while drawing in the \textit{TE-Pose} condition. However, in terms of visual feedback quality and overall performance, \textit{AU-Pose} is expected to be superior to the other two conditions since autonomous camera pose generation can enhance visual feedback quality, thereby aiding the drawing process.

$\boldsymbol{H_1}$: we predict that participants would experience higher mental and physical demand in \textit{TE-Pose} compared to \textit{AU-Pose} since the user needs to control the camera robot while finishing the drawing task which creates extra burden. 

\subsection{Analysis Metrics}

We quantified participants’ experiment completion times for each condition. The drawn results were also scanned to compare with the reference shape to evaluate the drawing accuracy based on Hu Moments \cite{hu1962visual} which calculates the distance between two images. Note that a higher distance value indicates larger shape differences. Further, we also analyzed participants' ratings from the post-experiment questionnaire for certain properties defined in Fig. \ref{fig:boxchart}. 






\vspace{-2pt}

\section{Results and Discussion}

\vspace{-2pt}

\begin{figure}
    \centering
    \subfloat[Viewpoint quality \label{fig:AVQ}]{\includegraphics[width=0.24\textwidth]{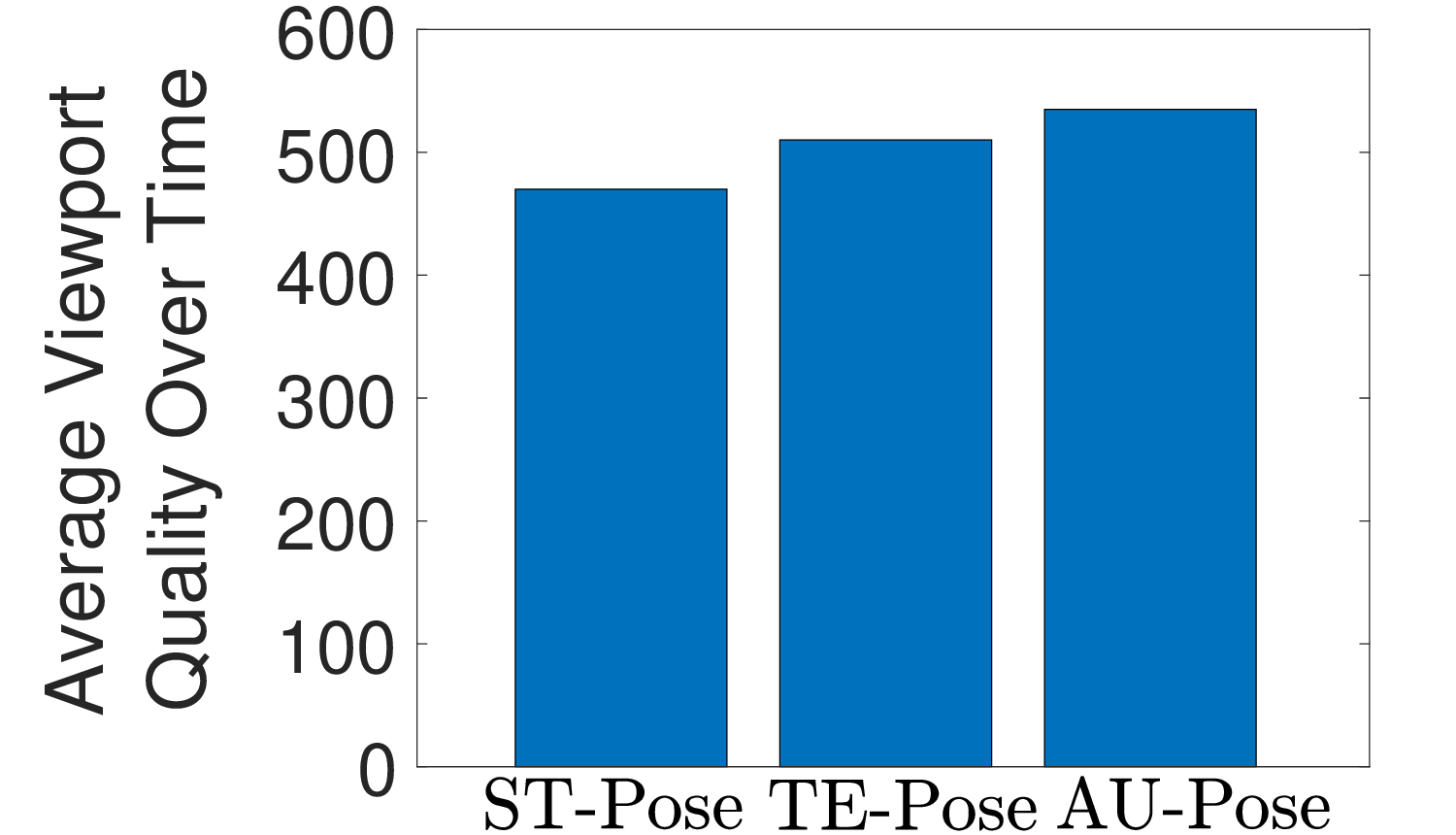}}
    \subfloat[Shape similarity metric \label{fig:Distance}]{\includegraphics[width=0.23\textwidth]{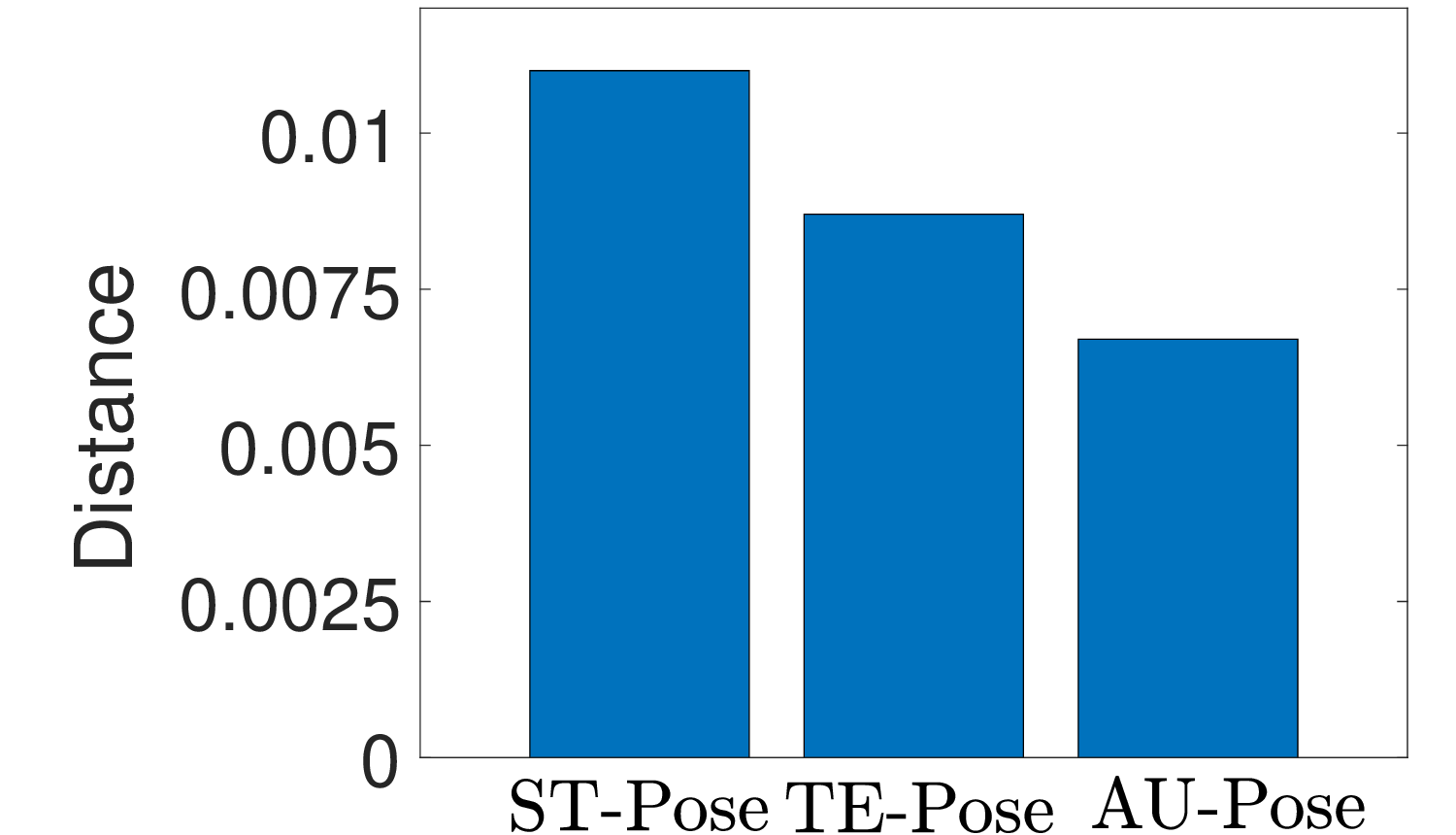}}
    \\[-10pt]
    \subfloat[Completion times \label{fig:boxchart_completiontime}]{\includegraphics[width=0.24\textwidth]{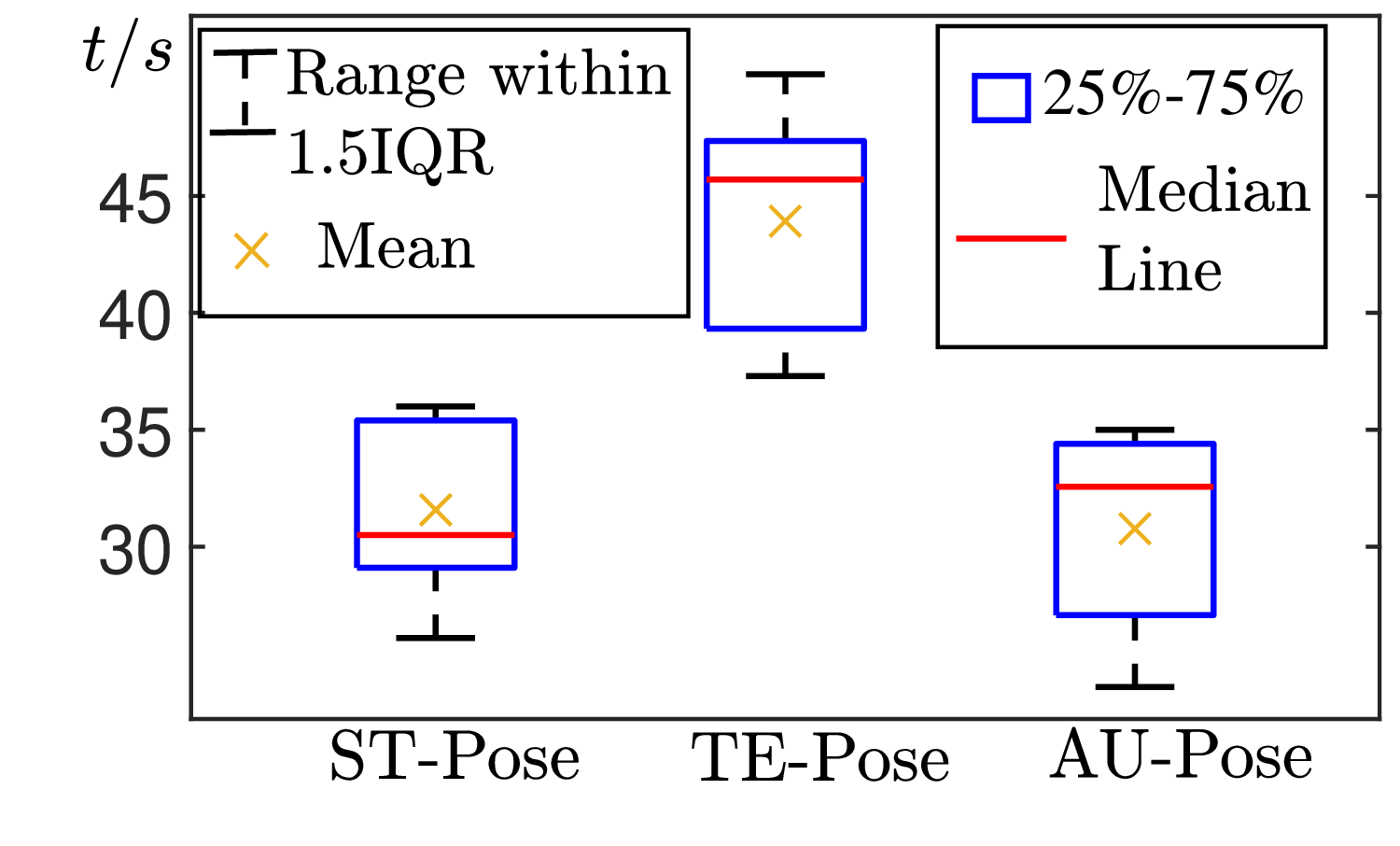}}
    \subfloat[Scanned drawn shapes\label{fig:Scanned_shapes}]{\includegraphics[width=0.22\textwidth]{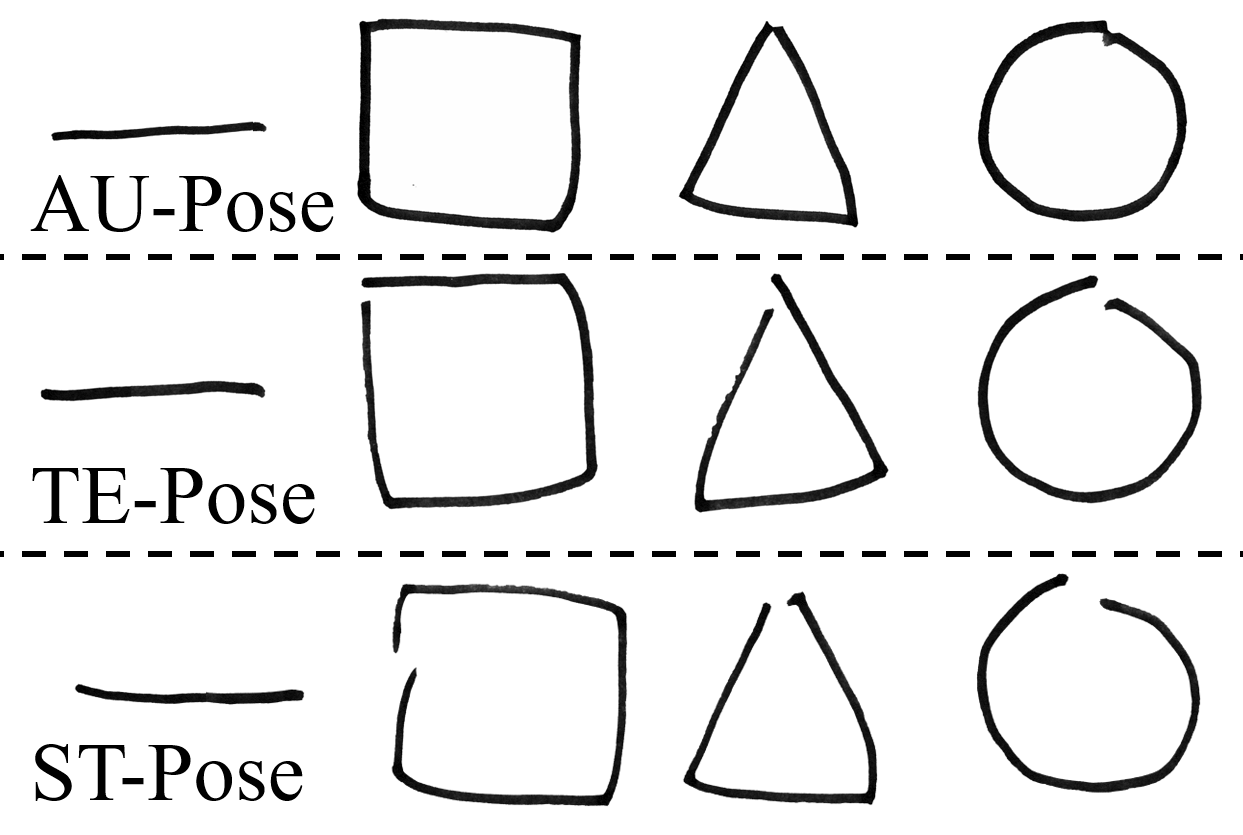}}
    \caption{Experiment Result.}
    \vspace{-20pt}
\end{figure}

Averaged viewpoint quality metric results among trials for each condition are shown in Fig. \ref{fig:AVQ}. The proposed \textit{AU-Pose} can deliver a better overall visual feedback quality than \textit{ST-Pose} and \textit{TE-Pose} based on the defined viewpoint quality metric which partially support hypotheses $\boldsymbol{H_0}$. 


Completion times are displayed in Fig. \ref{fig:boxchart_completiontime}, which demonstrates a similar completion time between \textit{AU-Pose} and \textit{ST-Pose}, but shorter than \textit{TE-Pose}, thereby confirming our hypothesis $\boldsymbol{H_0}$. Indeed, the shorter and similar completion time in the \textit{AU-Pose} and \textit{ST-Pose} conditions is expected, as participants in both of these conditions did not have the additional task of manually teleoperating the camera as in the \textit{TE-Pose} condition.
 
Results of the averaged shape similarity metrics shown in Fig. \ref{fig:Distance} demonstrate the \textit{AU-Pose} also performs better than \textit{ST-Pose} and \textit{TE-Pose}. A sample of scanned drawn shapes is shown in Fig. \ref{fig:Scanned_shapes} which indicates the participants completed the trials properly.

Results of the post-experiment questionnaire data are shown in Fig. \ref{fig:boxchart}. The results show the \textit{AU-Pose} requires similar mental and physical demands as the \textit{ST-Pose} while providing higher visual feedback quality. For the user-controlled camera robot method, even though the visual feedback quality is higher than the \textit{ST-Pose}, the mental and physical demands of the user are much higher which supports the hypotheses $\boldsymbol{H_1}$. The overall performance of the \textit{AU-Pose} is the best among the three conditions which supports $\boldsymbol{H_0}$.

	\section{Conclusions}
In this paper, we developed a drawing robot avatar that allows users to draw via telepresence. A novel method is proposed for automatically generating camera robot poses to acquire better visual feedback for the user to avoid occlusion. 
The evaluation of the complete telepresence system is done by both quantitative and qualitative measurements. 
Future work includes further experiments with complex shapes and evaluating the effectiveness of the introduced memory effect weight function. Testing can also be done in industrial and medical settings to further evaluate the proposed camera viewpoint automation method with existing ones.









\bibliographystyle{IEEEtran}
\bibliography{IEEEabrv,reference}

\end{document}